%% file: main.tex
\documentclass[journal]{IEEEtran}
\usepackage{mathptmx}
\usepackage{amsmath,amssymb,amsthm,bm}
\usepackage{graphicx,booktabs,multirow,array,tabularx,balance}
\usepackage{xcolor,cite,url}
\usepackage[hidelinks]{hyperref}

\newtheorem{proposition}{Proposition}
\DeclareMathOperator{\clip}{clip}
\newcommand{\widefig}[4]{\begin{figure*}[!t]\centering\includegraphics[width=#1\textwidth]{figs/#2.pdf}\caption{#3}\label{#4}\end{figure*}}
\makeatletter
\newcommand{\firstpagefigure}{\begin{minipage}{\textwidth}\def\@captype{figure}\centering
\includegraphics[width=.94\textwidth]{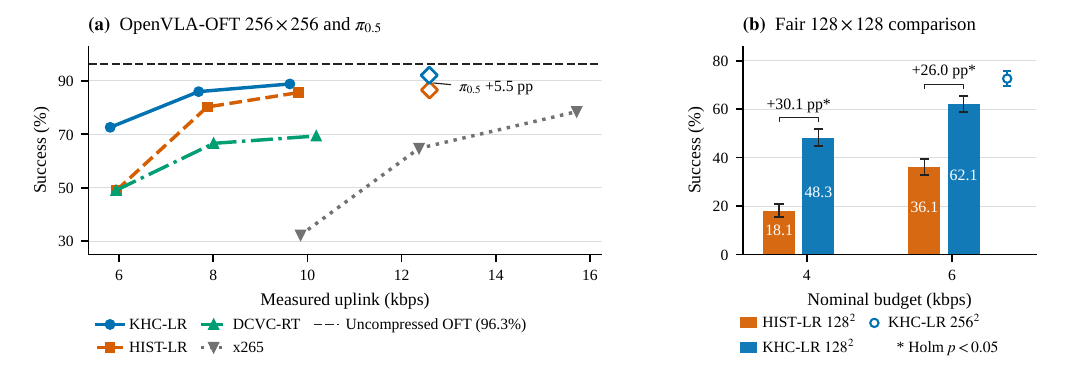}
\caption{Manipulation success versus uplink rate. (a) OpenVLA-OFT at $256\times256$ and the paired $\pi_{0.5}$ result at 12.59\,kbps. (b) Matched $128\times128$ streams at two budgets; the open circle gives the $256\times256$ KHC-LR result at 6\,kbps. Error bars are Wilson intervals.}\label{fig:teaser}\end{minipage}\par\medskip}
\makeatother
\begin{document}
\title{Kinematic Hypothesis Context: Embodiment-Conditioned Learned Video Coding for Bandwidth-Limited Robot Policy Uplinks}
\author{Yuze~Lu\\\textit{Tsinghua University, Beijing, China}}
\markboth{Kinematic Hypothesis Context for Robot Policy Uplinks}{Lu: Kinematic Hypothesis Context}
\IEEEaftertitletext{\firstpagefigure}
\maketitle
\raggedbottom
\input{sec_abstract}
\input{sec_intro}
\input{sec_related}
\input{sec_theory}
\input{sec_method}
\input{sec_setup}
\input{sec_results}
\input{sec_discussion}
\IEEEtriggercmd{\balance}
\IEEEtriggeratref{30}
\bibliographystyle{IEEEtran}
\bibliography{refs}
\end{document}

%% file: sec_abstract.tex
\begin{abstract}
A robot sends camera observations to a remote policy while also reporting the motion that produced them. We use the latter signal to reduce the cost of the former. Kinematic Hypothesis Context (KHC) lets both endpoints warp a decoded reference using joint states, link geometry and camera calibration. A learned codec fuses the hypotheses with image history; the geometry requires no separate image side stream, and transmitted joints count toward the uplink budget. Across nine measured rate points (0.0124--0.1261 pooled bits/pixel), the deployed model KHC-LR saves 27.6\% in PSNR BD-rate and 29.7\% at equal policy action deviation over a matched history codec. At equal action deviation, it saves 52.4\% over x264, 56.2\% over SVT-AV1, 30.1\% over DCVC-RT, and 5.1\% over VTM-23 (12.0\% in the narrowband range). Across 48,000 manipulation rollouts with two frozen policies, matching both codecs at $128\times128$ yields gains of 30.13 and 26.00 percentage points at 4 and 6\,kbps budgets; with the flow-matching policy $\pi_{0.5}$ the gain is 5.50 points at a measured 12.59\,kbps. A batched implementation halves dual-camera transmit latency in a matched profile. Real-robot transfer and delay-aware control identify two separate deployment requirements: informative calibration motion and query turnaround that fits the action horizon.
\end{abstract}

%% file: sec_intro.tex
\section{Introduction}\label{sec:intro}
\IEEEPARstart{A}{ wrist} camera moves because the robot moved. A remote manipulation policy ordinarily receives a compressed image of that motion and a separate, accurate measurement of the joints that caused it. An image-history codec treats the two streams as unrelated: it spends bits on the displacement of the gripper or the work surface even when the receiver could have predicted much of that displacement. This redundancy matters when two cameras must share a link of only a few kilobits per second. It also suggests an alternative to lowering the image resolution: keep the pixels, but supply the predictor with their known geometry.

Consider a gripper passing in front of a stationary object. In the fixed camera, a link changes position and exposes part of the table. A receiver can move the previously decoded link appearance with its forward kinematics and retrieve previously seen background. In the wrist camera, by contrast, the entire work surface moves across the sensor. With the calibrated camera motion, a pixel on a static surface moves along a one-dimensional inverse-depth family. Neither operation reveals the hidden face of a newly introduced object. A useful codec must exploit the first two sources of predictability while continuing to transmit evidence for the third.

We call this representation \emph{Kinematic Hypothesis Context} (KHC). At both endpoints it constructs candidate images from a decoded reference, measured joints and a shared geometric model. A learned fusion network supplies the candidates to a predictive codec rather than prescribing a warp for every pixel. The deployed KHC-LR model uses nine rate points, including two low-rate points learned with the same recipe as its history-only control (HIST-LR). Image bits, headers and state bytes all count. A separate seven-rate model, KHC-Base, permits a controlled comparison of predictive coding and frame replacement at the same encoder settings.

Figure~\ref{fig:teaser} illustrates why a transport result should not be summarized by one distortion number. At a 4\,kbps budget, identically downsampled inputs yield 386 successful trials with KHC-LR and 145 with HIST-LR out of the same 800 initial conditions. Conversely, if a remote controller waits for each packet and action block, even an accurate image can arrive too late. We therefore evaluate three distinct outcomes: the bytes required for a reconstructed image, the action requested by a frozen policy on that image, and the success of the resulting sequence of actions.

This paper makes four contributions.
\begin{enumerate}
\item We formulate robot video as conditional source coding and identify the motion fields a global displacement cannot represent. A regional analysis of 326,592 coded model frames separates savings on robot pixels and geometrically overlapping background from the cost of newly visible content.
\item We develop a shared multi-hypothesis predictor, variable-rate conditional host and arithmetic-byte controller. In a two-camera, 32-plane profile, batching reduces median transmit time from 229.4 to 114.5\,ms; on the deployed weights the full-range PSNR BD-rate difference between implementations is $+0.00029\%$.
\item We evaluate six standard or learned anchors and two policies. KHC-LR needs 5.10\% fewer bits than VTM-23 at equal action deviation, despite requiring more bits at equal PSNR. Its closed-loop advantage persists when both methods reduce resolution and when the policy is changed to $\pi_{0.5}$.
\item We separate the benefit of predictive coding from the synthesis used on a skipped frame and identify a timing limit in 48,000 closed-loop rollouts. Tests on two real robots distinguish adaptation in a target domain from zero-shot transfer with weak calibration motion.
\end{enumerate}

%% file: sec_related.tex
\section{Related Work}\label{sec:related}
\subsection{Video Coding with Context}
AVC, HEVC, VVC and AV1 exploit motion-compensated prediction, block partitioning and transform coding~\cite{wiegand2003h264,sullivan2012hevc,bross2021vvc,han2021av1}. Their motion models are inferred from the video; the decoder generally receives a description of how to predict the next block. Multiview extensions exploit inter-camera prediction~\cite{tech2016mvhevc}. Our question is different: can a receiver derive a useful temporal prediction from the joints of the device carrying the cameras, while continuing to code each view causally?

Learned codecs can encode optical flow and residuals~\cite{lu2019dvc,agustsson2020ssf}, predict features~\cite{hu2021fvc}, or condition analysis, synthesis and entropy probabilities on temporal context~\cite{li2021dcvc,sheng2022tcm,mentzer2022vct}. Scale hyperpriors and hierarchical priors provide rate models for the remaining latents~\cite{balle2018hyperprior,minnen2018joint}. KHC changes what both ends know before they code a frame: it inserts an externally determined family of correspondences into the temporal context. We compare against the official real-time DCVC-RT and diverse-context DCVC-DC models~\cite{jia2025dcvcrt,li2023dcvcdc}. The former overlaps the low-rate regime; the measured grid of the latter starts at 0.0614 bits/pixel. This is a property of the tested operating points, not a lower bound on the codec family.

\subsection{Coding for Machines and Geometric Prediction}
Classical conditional and distributed coding formalize the value of correlated side information~\cite{gray1972conditional,slepian1973,wyner1976,girod2005dvc}. Learned decoder-side information can likewise reduce rate~\cite{ayzik2020dsi}. Plane-sweep reconstruction enumerates candidate depths under known camera motion~\cite{collins1996planesweep}. We use such planes as codec features and combine them with hypotheses for independently moving links, an end-effector plane and a background store. Neither view uses an uncoded side image or a scene-depth measurement at the receiver.

Video coding for machines can preserve downstream decisions rather than maximize fidelity to all source pixels~\cite{duan2020vcm,shao2022taskoriented}. Ge et al.~\cite{ge2024taskencoder} adapt encoder control to a machine task; Gadot et al.~\cite{gadot2025rldot} learn block-level coding decisions from task feedback. KHC instead uses robot geometry available before the next action, without optimizing the frozen policy or using task rewards to choose packets. Measuring both image distortion and action deviation makes the distinction experimentally visible.

\subsection{Remote Policies and Action Horizons}
Vision--language--action (VLA) policies amortize inference over short action sequences~\cite{kim2024openvla,kim2025oft,zhao2023act}. The flow-matching $\pi_0$ family uses a different action-generation mechanism~\cite{black2024pi0,pi05}; it provides a test of whether a coding change is specific to one visual-action interface. Cloud and fog robotics separate sensing from compute~\cite{kehoe2015cloud,ichnowski2023fogros2}, but compression alone does not guarantee a timely response. Real-time chunking overlaps inference with execution of a previous chunk~\cite{black2025rtc}. We measure synchronous and delay-aware stop-and-wait operation separately, and relate the resulting action starvation to the time spanned by a chunk.

%% file: sec_theory.tex
\section{What Shared Kinematics Can Predict}\label{sec:theory}
Let $X_t$ denote a camera image, $C_s$ the receiver's decoded history, and $Q=(q_s,q_t,\mathcal M,\Theta)$ the joint states, robot model and calibration. The index $s=t-k$ denotes the reference actually used, not necessarily the previous control step. All statements below concern the image stream conditioned on an available $Q$. Transmitting seven additional joint angles, when needed, contributes to the measured total uplink.

\begin{proposition}[Conditional coding]\label{prop:rate}
For a common distortion measure and admissible reconstruction alphabet, $R_{X|C,Q}(d)\le R_{X|C}(d)$. For fixed discrete symbols $Z$ the ideal entropy-rate saving is exactly $I(Z;Q\mid C)$, strictly positive when the two are conditionally dependent.
\end{proposition}
The first statement follows by allowing the conditional test channel to ignore $Q$~\cite{gray1972conditional,cover2006}; the second follows from
\begin{equation}
 H(Z\mid C)-H(Z\mid C,Q)=I(Z;Q\mid C).
 \label{eq:info}
\end{equation}
The distinction matters. A shared signal that predicts a quantized residual can lower its code length. It need not strictly improve every point of a lossy rate--distortion curve; for example, both rates are zero at a sufficiently large distortion. Measured packet savings also include the consequences of imperfect entropy models and finite headers.

\begin{proposition}[Rotation is not a translation]\label{prop:rot}
On a finite image patch let $v(u)$ be the first-order displacement of pixel $u$. Any single integer translation $\delta\in\mathbb Z^2$ has mean squared displacement error at least $\mathbb E\|v-\mathbb Ev\|^2$. Under a small rotation of size $\epsilon$, this term is of order $\epsilon^2$ when the first-order field has nonzero spatial variance.
\end{proposition}
For normalized coordinates $(x,y)$, an active rotation $\omega$ of a camera-frame point induces
\begin{equation}
 v_\omega=
 \begin{pmatrix}
 -xy\omega_x+(1+x^2)\omega_y-y\omega_z\\
 -(1+y^2)\omega_x+xy\omega_y+x\omega_z
 \end{pmatrix}.
 \label{eq:flow}
\end{equation}
This field depends on location. An additive broadcast of a global pose vector is also spatially constant before it interacts with image features. Neither a single translation nor this additive term specifies the projective field. A nonlinear network with pose, image features and wide spatial support is not subject to that expressivity claim. Nor is displacement error necessarily visible on a textureless patch. GVEC tests what a broadcast vector buys in a specific matched network.

\begin{proposition}[Depth-indexed prediction]\label{prop:depth}
Let $(R,t)$ map the target camera into the reference camera, $p=K_t^{-1}\tilde u$ with $p_3=1$, and $\rho=1/Z$. A visible static point projects to
\begin{equation}
 u_s(\rho)=\pi\!\left(K_s(Rp+\rho t)\right),\quad
 H(\rho)=K_s(R+\rho t e_3^\top)K_t^{-1}.
 \label{eq:plane}
\end{equation}
For a uniform inverse-depth grid, the nearest plane incurs displacement no greater than $\tfrac12\Delta\rho\sup_\rho\|\partial u_s/\partial\rho\|$ on an interval of positive projective depth.
\end{proposition}
Pose reduces the search over general two-dimensional motion to a one-dimensional family. Pure rotation removes the depth dependence; translation makes plane spacing matter. Large stride can increase both the utility of pose and the amount of previously invisible content. A new pixel outside the reference field cannot be recovered by increasing the number of planes.

\begin{proposition}[From distortion to control]\label{prop:task}
Let $S_A(R)$ be task success for codec $A$ at rate $R$. If codec $B$ attains the same task-relevant quality at a fraction $1+\gamma$ of the rate and the observation timing is unchanged, then the local rate-shift model gives
\begin{equation}
 S_B(R)-S_A(R)\simeq
 -\frac{\partial S_A}{\partial\log R}\log(1+\gamma).
 \label{eq:task}
\end{equation}
\end{proposition}
For a fixed policy state, a first-order analysis weights each pixel error $e_u$ by its action Jacobian: $\|\Delta a\|\le\sum_u\|J_u\|\|e_u\|+O(\|e\|^2)$. Thus a change at the gripper may matter more than the same error on distant background. When a packet is skipped the error includes prediction of the current view from an older reference; when delivery is delayed the policy also acts on an older observation. Neither effect is captured by substituting a global PSNR BD-rate into~\eqref{eq:task}.

\begin{proposition}[Sensitivity to hand--eye calibration]\label{prop:cal}
Write the relative end-effector pose as $T_\Delta$, the hand--eye calibration as $E$, the camera-frame motion as $T_c=E^{-1}T_\Delta E=I+\tau_c$, and a perturbed calibration as $E'=E(I+\varepsilon)$. With $[A,B]=AB-BA$,
\begin{equation}
 E'^{-1}T_\Delta E'-T_c=(I+\varepsilon)^{-1}[\tau_c,\varepsilon].
 \label{eq:comm}
\end{equation}
\end{proposition}
Error is governed by the product of motion and calibration perturbation. Small inter-frame motion can therefore conceal a poor calibration. In contrast, a textured boundary crossing a reference field can lose predictive value quickly even at correct calibration. Supplementary Material, Sec.~S-I, derives the bounds and gives an explicit texture-correlation model for this latter effect.

%% file: sec_method.tex
\section{Kinematic Hypothesis Context}\label{sec:method}
\subsection{A Shared Predictor at the Two Endpoints}
A fixed camera observes the robot and workspace; a second camera is mounted on its wrist. On a query, the robot has both originals $x_t^{(c)}$. The remote policy has neither. The shared information is a calibrated camera/robot model, joint telemetry and the decoded reference $\hat x_s^{(c)}$ held by each endpoint. At each P query both sides reconstruct the same candidate geometry from these inputs. Only the sender sees the original when deciding what to code. Figure~\ref{fig:overview} separates this local computation from signals that cross the link.

\widefig{1}{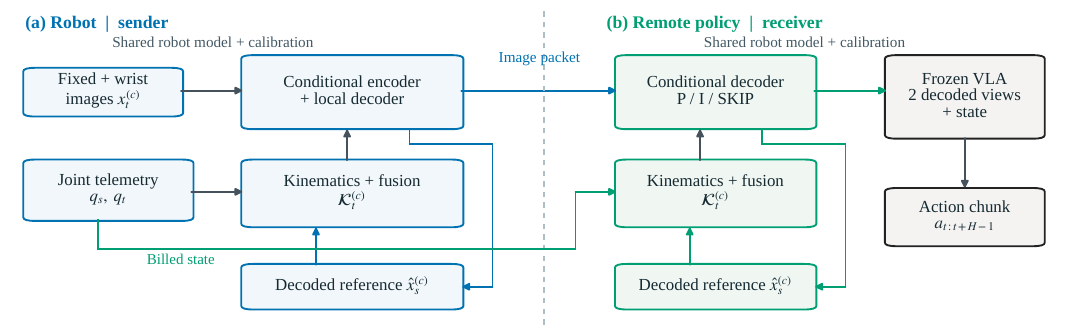}{Robot and policy server maintain separate reference buffers. Each feeds its locally decoded frame back into the next prediction. The image packet carries coded streams or a two-byte SKIP record; the parallel telemetry path is billed. The two reconstructed views and state enter a single policy.}{fig:overview}

\subsection{Hypotheses for Static Surfaces and Moving Links}
For each camera there are $J=D+5$ slots, indexed $j=0,\ldots,D+4$. Each contributes a warped RGB image and a validity map. Slot 0 is the unwarped reference. Slots $1,\ldots,D$ implement~\eqref{eq:plane} at uniformly spaced inverse depths between $0.3333$ and $12.5\,\mathrm{m}^{-1}$. The fixed camera has little background ego-motion; the wrist camera requires depth-dependent reprojection across much of its field.

Slot $D+1$ is the robot composite. Given rendered depth and link identifier $\ell$ at the current pixel, the corresponding world point is carried back with that link to the reference pose. If $T_{ab}$ maps coordinates in $b$ into $a$, the backward map is
\begin{equation}
 u_s=\pi\!\left(K_s[I_3\;0] T_{c_s w} T_{w\ell}(q_s)
 T_{\ell w}(q_t) T_{wc_t}\,[Z_tK_t^{-1}\tilde u;1]\right),
 \label{eq:robot}
\end{equation}
where the camera projection uses the first three Cartesian coordinates. A depth disagreement exceeding 5\% of depth plus 1\,cm flags a self-occlusion. Slot $D+2$ is a plane through and attached to the end effector; it offers a candidate for transported objects without requiring object pose. Slot $D+3$ carries background memory under the fixed camera, and $D+4$ is an auxiliary slot, normally identity. No scene depth enters the deployed predictor. An oracle that uses scene depth occupies the auxiliary slot only in an ablation.

\widefig{1}{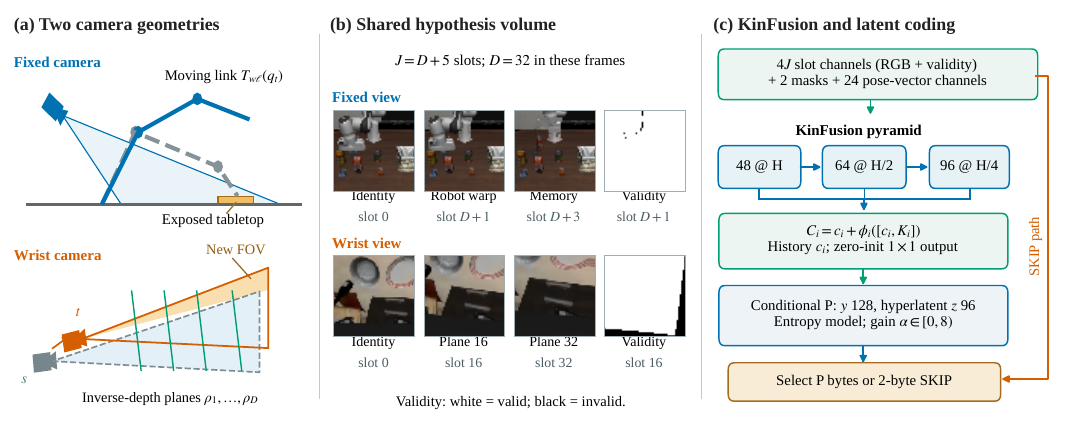}{(a) A moving link exposes tabletop pixels to the fixed camera; wrist motion changes the field of view across the depth planes. (b) Stored RGB hypotheses and binary validity maps at stride eight. For the wrist's 16th depth plane, black marks reference samples outside its valid support. (c) KinFusion injects three scales into the history context; an indexed SKIP bypasses P-frame coding.}{fig:method}

\subsection{Conditioning a Low-Delay Latent Codec}
The $4J$ slot channels are concatenated with two global mask maps and 24 broadcast channels. The global vector is used by the pose-vector control; it is not a surrogate for the spatial warps. KinFusion encodes the volume into three feature scales. Each modifies the matched history context $c_i$ through a $1\times1$ mixer $\phi_i$ with zero-initialized final layer:
\begin{equation}
 C_i=c_i+\phi_i([c_i,K_i]),\quad
 p(\hat y\mid\hat z,C_3),\qquad i=1,2,3 .
\end{equation}
The main host is a DCVC-style conditional latent codec with a coded hyperprior $\hat z$, not an optical-flow estimator. The contexts enter analysis, entropy prediction and synthesis. HIST uses the same input dimensions and parameter budget but repeats the decoded reference. HIST2 supplies a second reference, GVEC supplies the broadcast pose, and SHUF uses a mismatched episode's geometry. These controls distinguish transport of appearance from simply adding inputs.

\subsection{One Model for Nine Operating Points}
For four-frame clips the objective is expected P-frame bits per pixel plus $\lambda\,255^2\operatorname{MSE}$, using the codec's own previous reconstruction. Learned per-channel gain units vary continuously with $\alpha$; gain and inverse-gain magnitudes are geometrically interpolated between adjacent rows. The seven-row backbone and matched history backbone are trained separately. For each, two lower-rate rows are initialized by extrapolation from its lowest rows, followed by an identical 50,000-update fine-tune. In increasing order the nine distortion multipliers are
\begin{align}
 10^{-4}\!\times\!(&0.516,\ 1.14,\ 2.5,\ 5.5,\ 12,\nonumber\\[-.15em]
                    &27,\ 59,\ 130,\ 280).
\end{align}
The corresponding frozen intra qualities are $(1,1,1,1,2,3,4,5,6)$~\cite{minnen2018joint}. Thus low-rate extension changes the P-frame operating points, not the first-frame bitrate at quality~1. In training $\alpha$ is drawn continuously; on the wire it is quantized in units of $1/32$, with the largest representable value $7.96875$. We use the final, not a success-selected, checkpoint.

\subsection{Rate Control and Kinematic SKIP}
The sender compares the actual P payload against a two-byte alternative that names a shared hypothesis:
\begin{align}
 J_{\rm P}&=8|\mathrm{payload}|/(HW)+\lambda_e\,255^2d_{\rm P},\\
 J_{\rm skip}&=16/(HW)+\lambda_e\,255^2\min_j d(x_t,\tilde x_j).
\end{align}
Here $d$ is per-pixel MSE and $H,W$ denote image dimensions in these two expressions. SKIP wins only at a strictly lower cost. The receiver displays that candidate, and both buffers update to it. HIST-SKIP repeats its reference; KHC can instead show a link at its newly reported pose. A lost packet and a deliberate SKIP are distinct events.

At each query the controller deducts an exponentially averaged state and header allowance from the per-query byte budget and shares the remainder across the two views, with a per-camera floor $B^\star\ge4$ bytes. For actual camera bytes $b$ it updates
\begin{align}
 \beta'&=\clip(\beta+b-B^\star,-32 B^\star,32 B^\star),\nonumber\\
 e&=\clip((b-B^\star)/B^\star,-1,1),\nonumber\\
 \alpha'&=\clip\!\left(\alpha-0.3e-0.6\,\beta'/(8B^\star),-8,R-1\right).
 \label{eq:bucket}
\end{align}
Here $R$ is the number of trained rate points. Negative $\alpha$ codes a P frame at row zero but sets $\lambda_e=\lambda(0)4^\alpha$, favoring cheaper SKIPs. The integral term acts on arithmetic bytes; averaging logarithms would undercount occasional large packets. Each uplink includes state, view headers, stream lengths and coded bytes. The policy state is transmitted losslessly by XOR prediction of float words; the seven additional joint angles used by KHC are included under the same accounting.

\subsection{Photometric Self-Calibration}
For a wrist camera without trusted extrinsics, the fitted variables are selected components of hand--eye pose and camera intrinsics. A robust soft minimum over depth-plane photometric discrepancies provides a gradient without measured scene depth. Samples outside the image or with insufficient valid planes do not contribute. In simulation a robot mask can exclude moving pixels; on DROID and UR5 no rendered robot-depth mask is available, so all geometrically valid pixels can contribute, including robot pixels. A moving hand, a textureless wall or small rotations can make a low reprojection loss consistent with the wrong geometry. Target-domain calibration and zero-shot deployment are evaluated separately.

\subsection{Batched Projection}
The reference path constructs one sampling grid per depth plane. The fast path stacks the homographies and grids, reshapes the batch and plane dimensions, and applies bilinear resampling to the entire stack. It uses the same inverse depths, pixel centers, visibility tests and interpolation convention. This transformation removes repeated kernel launches; it does not change the packet syntax. Floating-point reductions can still differ, so we compare bytes and reconstructions within each encoder--decoder path and compare rate--distortion curves across paths. Section~\ref{sec:systems} reports the geometry and end-to-end timings separately.

%% file: sec_setup.tex
\section{Experimental Design}\label{sec:setup}
\subsection{Scenes, Cameras and Splits}
LIBERO~\cite{liu2023libero} supplies tasks, demonstrations and initial conditions; MuJoCo and robosuite~\cite{todorov2012mujoco,zhu2020robosuite} provide synchronized cameras and robot geometry. From LIBERO-90 we use 810 demonstrations for training and 90 for validation, one held-out validation trajectory per task. The offline test comprises five demonstrations for each of 40 different tasks from Spatial, Object, Goal and Long (LIBERO-10): 200 trajectories, none used for model selection. We render both views at $256\times256$. For the spatial-resolution comparison both codecs transmit $128\times128$ images and use the same resizing path to the policy.

DROID~\cite{khazatsky2024droid} provides a different Franka environment, a wrist view and two external views; 63 training, seven validation and 30 test trajectories are split by laboratory. We fit wrist geometry on training data, select the rotation-only candidate on validation data, and adapt both deployed codecs for 100,000 updates. Both focal lengths remain at 130 pixels. The Berkeley Autolab UR5 experiment~\cite{berkeleyUR5} instead freezes all network weights: 20 calibration trajectories are disjoint from 80 test trajectories. The UR5 camera frames are $640\times480$. Both real datasets are evaluated offline without robot-only rendering.

\subsection{Training and Anchors}
The conditional hosts have 7.0 million parameters, with 48/64/96-channel contexts, 128 main latent channels and 96 hyperlatent channels. Base models use 200,000 AdamW updates, a $10^{-4}$ learning rate and batch eight. Each lower-rate extension uses batch eight, seed zero, 1,000 warm-up updates and a cosine schedule from $5\times10^{-5}$. Clips contain four frames with strides drawn from $\{1,2,4,8\}$. The loss, data, gain-table extension and update count are identical for KHC-LR and HIST-LR. The nine ablation variants and their history anchor each use a separate 100,000-update training run.

Four standard anchors use low-delay x264 (4:2:0), x265 (4:4:4), SVT-AV1 (4:2:0), and VTM-23 (4:2:0); VTM 4:4:4 is an additional chroma-format comparison. Two learned anchors use the official DCVC-RT and DCVC-DC weights. Inputs, cadence and intra refresh (32 coded frames) are shared for the primary LIBERO offline comparison. Offline rates count coded bytes with camera-stream framing. The regional analysis assigns estimated symbol information to spatial masks.

\subsection{From Sensor Samples to Decisions}
Offline tests have three measurement levels. (i) At the image level we report RGB PSNR, MS-SSIM~\cite{wang2003msssim} and AlexNet LPIPS v0.1~\cite{zhang2018lpips}. (ii) At the policy level, both decoded views, the original proprioception and instruction are supplied to a frozen policy. The mean absolute difference between its $8\times7$ decoded-view and original-view action chunks gives action L1. This score uses both views jointly. (iii) In closed loop the policy acts, altering subsequent observations. The paired outcome is task success. The seven- and nine-rate offline operating curves contribute 61,964 and 79,668 frame--rate observations, respectively; the same frames at different rates are not independent samples.

A budget is an allowed total uplink rate; a measured rate is $8\sum b/(1000\,T)$ on the control timeline. At 20\,Hz with stride eight there are 2.5 two-view queries per second. For the deployed codec, the nine \emph{pooled test} image operating points cover 0.0124--0.1261\,bpp, or 4.06--41.33\,kbps of image payload at that cadence; closed-loop telemetry, framing and the SKIP decisions change the transmitted rate. Frame distortion and byte rates are averaged with frame and pixel counts, respectively.

\subsection{Paired Rollouts and Inference}
We evaluate frozen OpenVLA-OFT~\cite{kim2025oft} (seven-billion-parameter family, eight-step action chunks) and the flow-matching policy $\pi_{0.5}$~\cite{pi05}. Each paired comparison fixes the suite, task, initial state and seed. The 48,000 formal episodes consist of 20,800 base-model rollouts, 11,200 extended OpenVLA controls, 1,600 earlier second-policy rollouts, 4,800 lower-rate rollouts, and 9,600 additional trials (3,200 matched-resolution, 1,600 lower-rate second-policy, and 4,800 delay-aware). Pilot episodes used to locate the steep part of the success curve are not counted. Reused anchor episodes contribute once to the inventory, regardless of how many contrasts use them.

Ordinary rollouts advance with a fresh decoded observation at each scheduled query; their rate uses logical control time, not wall-clock execution. A second set explicitly serializes packet bytes, applies measured encoding, decoding and policy compute, adds a 40\,ms round trip, and accounts for action starvation step by step.

BD-rate~\cite{bjontegaard2001} integrates log rate over the two measured quality curves' shared interval. Lower is better for the test codec; LPIPS and action error are sign-reversed for integration. The narrowband analysis truncates the quality interval at the test codec's quality at 0.041\,bpp. The full anchor matrix uses 1,000 paired episode bootstrap draws; intervals and numbers of valid draws are given in Supplementary Material, Sec.~S-II. Single success proportions use Wilson intervals~\cite{wilson1927}. Differences use 10,000 task-cluster bootstrap draws (40 tasks), exact two-sided McNemar tests on discordant pairs~\cite{mcnemar1947}, and Holm correction within the specified comparison families~\cite{holm1979}.

%% file: sec_results.tex
\section{Results}\label{sec:results}
\subsection{Rate--Distortion and Action Fidelity}\label{sec:rd}
Table~\ref{tab:bdmain} and Fig.~\ref{fig:rd} compare image quality and policy action fidelity on the same held-out demonstrations. Against the matched HIST-LR codec, KHC-LR reduces full-range rate by 27.61\% for equal PSNR (95\% CI 26.71--28.56\% savings), 23.31\% for MS-SSIM, 28.67\% for LPIPS and 29.68\% for action L1. The low-rate PSNR saving is 28.74\%. This comparison holds host capacity, intra codec, training schedule and rate range fixed.

\input{tables/t_bd_main}
\widefig{1}{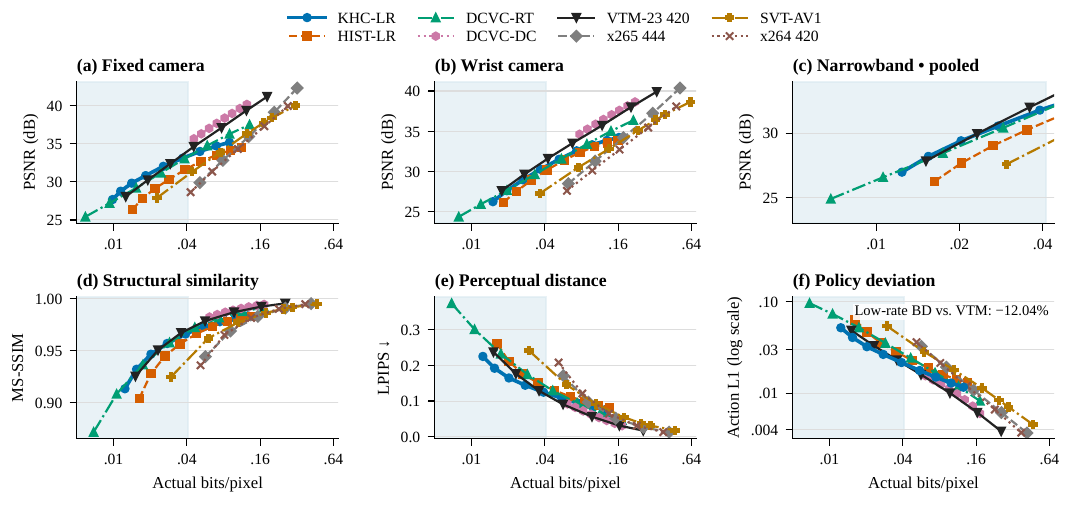}{Stride-eight operating points. Shading marks the narrowband region ($\leq0.041$\,bpp). Panels (a,b) separate the two views; (c)--(e) pool frames. Panel (f) uses joint-view policy action deviation on a logarithmic scale. KHC-LR's low-rate action BD-rate versus VTM-23 4:2:0 is $-12.04\%$.}{fig:rd}

The ranking changes with the target. VTM-23 4:2:0 requires fewer bits at equal full-range RGB PSNR: KHC-LR's BD-rate is $+16.13\%$. For equal action deviation the sign reverses: $-5.10\%$ (CI $[-8.83,-0.67]\%$) over the full interval and $-12.04\%$ ($[-15.25,-8.35]\%$) over the low interval. Against DCVC-RT the full PSNR difference is $+0.11\%$ ($[-6.87,7.86]\%$), but LPIPS and action BD-rates are $-24.74\%$ and $-30.14\%$. The action savings against x264, x265 and SVT-AV1 are 52.44\%, 50.67\% and 56.15\%, respectively. A small RGB improvement on background and a small improvement at a grasp boundary do not carry the same action weight.

The official DCVC-DC points start at 0.0614\,bpp, above the 0.041\,bpp narrowband limit. Its full-range PSNR values also do not overlap KHC-LR's combined-camera quality range. On their supported high-quality intervals KHC-LR costs 86.90\% more at equal MS-SSIM, 40.46\% more at equal LPIPS and 22.48\% more at equal action L1. Supplementary Material, Sec.~S-II, lists the common quality intervals.

\widefig{1}{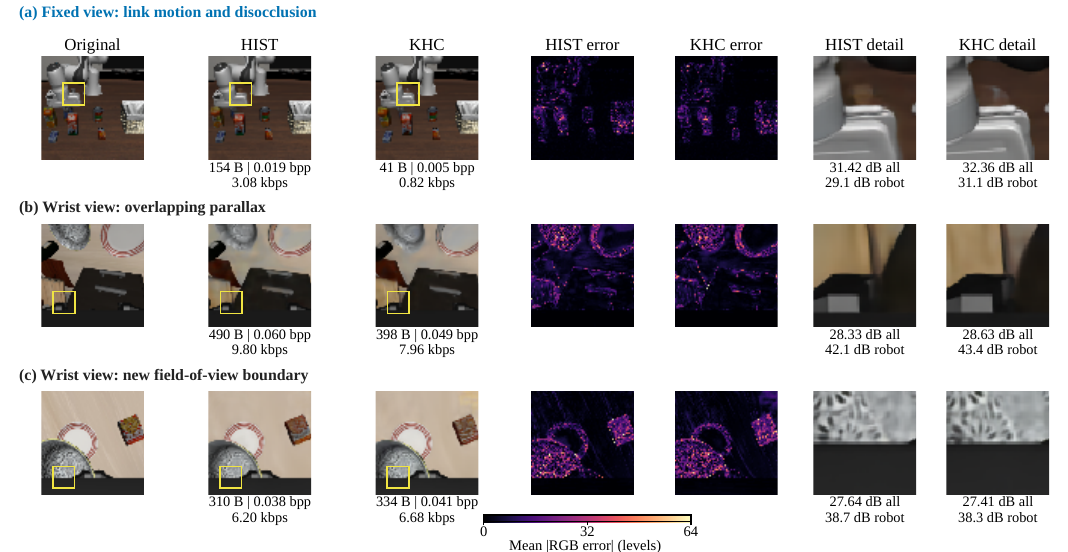}{Three coded frames at stride eight. Columns show the original, two reconstructions, mean absolute RGB errors (shared 0--64 intensity scale) and enlarged yellow boxes. Rates are per view at 2.5 queries/s; the detail labels give whole-frame and robot-mask PSNR.}{fig:qualcoding}

The three frames in Fig.~\ref{fig:qualcoding} show when a decoded reference remains useful. With the fixed camera, link motion exposes part of the tabletop; KHC uses 41 rather than 154 bytes and improves robot-region PSNR. In the overlapping wrist view, KHC uses 398 versus 490 bytes ($18.8\%$ fewer), while whole-frame PSNR rises from 28.33 to 28.63\,dB. At the larger new-field boundary the reference covers less of the wrist image: KHC uses 334 versus 310 bytes. Across hosts, conditional prediction saves 7.04\% versus history in a scale-space-flow codec; an entropy-only variant saves 10.03\% at unchanged reconstruction.

\subsection{Where Kinematic Prediction Saves Bits}\label{sec:mechanism}
Sparse transmission changes \emph{both} the reference displacement and the fraction of the current image covered by the reference. Across transmission strides 1, 2, 4, 8 and 16, pooled KHC-Base versus HIST PSNR BD-rates are $-28.60$, $-29.33$, $-27.67$, $-23.91$ and $-17.70\%$. A larger pose difference does not imply an increasing net saving. On the wrist, the spatially transported representation saves 11.14--18.49\% across four motion quartiles; giving the history network just a broadcast vector saves 1.13--2.56\%. On the fixed camera all four KHC savings exceed 37\%, while all four GVEC point estimates require more bits than HIST.

\widefig{1}{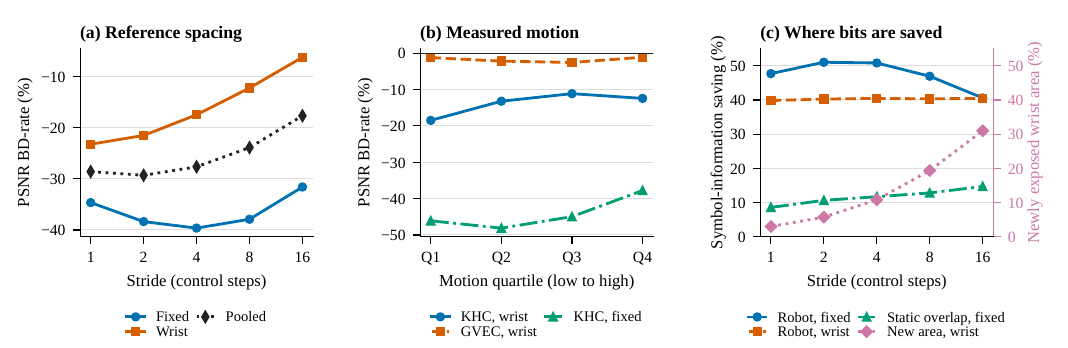}{(a) Base-model PSNR BD-rate against history at five strides. (b) Spatial transport and broadcast pose across motion quartiles. (c) Regional symbol-information saving (left axis) and newly exposed wrist area (right axis).}{fig:stride}

Figure~\ref{fig:stride}c locates this effect. The regional study comprises 80 held-out episodes, both cameras, three operating points and 326,592 coded model frames. Each regional ratio uses summed symbol information across its pixels. At the fixed camera, the robot-region saving is 47.71\% at stride one and 40.60\% at stride 16; the static, geometrically overlapping region increases from 8.55\% to 14.67\% in savings. On the wrist, robot-region savings stay near 40\%, but the fraction classified as newly visible or without reliable overlap grows from 2.97\% to 30.99\%. At stride 16 the wrist robot still saves 40.47\% while the whole-view symbol estimate \emph{increases} 6.02\%. These regional estimates exclude shared packet overhead.

\widefig{1}{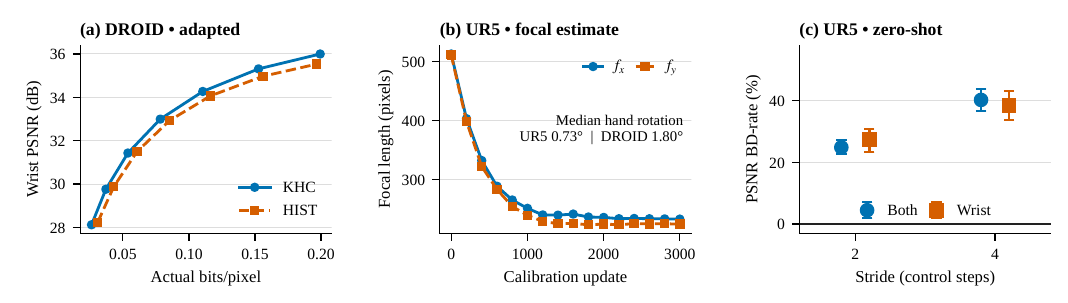}{Real-robot transfer. (a) DROID wrist test rates after target-domain adaptation. (b) UR5 focal lengths during the independent calibration fit; hand rotation has median $0.73^\circ$ and 90th percentile $4.48^\circ$. (c) Network-frozen UR5 PSNR BD-rate for the wrist alone and both cameras pooled, with paired episode intervals.}{fig:transfer}

\subsection{Matched-Budget Ablation and Robustness}\label{sec:ablation}
At 100,000 updates for every model, full KHC reduces rate by 25.67\% against HIST-100k (Table~\ref{tab:ablation}). Removing the robot composite leaves the smallest saving among the component removals, about 6\%; removing the planes leaves about 19\%. Dropping background memory or the end-effector plane leaves about 22\%. All component comparisons retain the same input dimensions. The oracle supplies measured scene depth to the auxiliary slot as an architectural upper bound.

\input{tables/t_ablation}
\widefig{1}{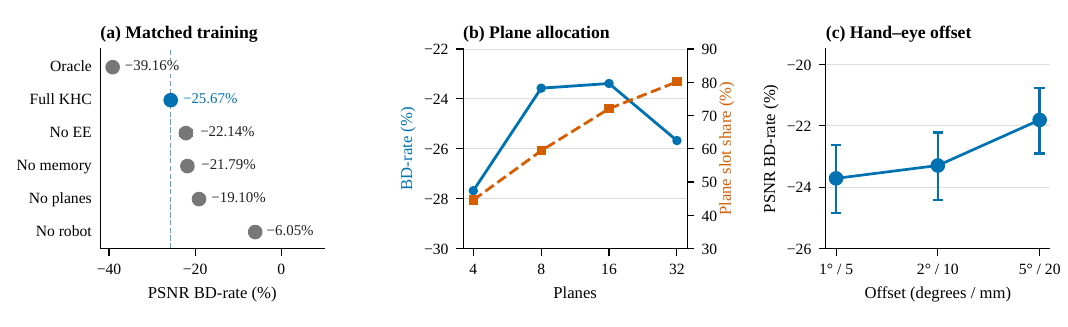}{(a) Component ablations versus HIST-100k; the dashed line marks full KHC. (b) Matched-budget BD-rate and the share of the $4J$ slot-channel $L_1$ norm assigned to depth planes. (c) Paired-episode intervals for simultaneous angular and translational offsets at both endpoints versus clean HIST.}{fig:ablation}

Four planes yield $-27.68\%$ BD-rate versus $-25.67\%$ for 32 planes. In a fixed-width fusion stem, the plane group accounts for 44.6\%, 59.4\%, 72.0\% and 80.2\% of the total \emph{slot} weight norm as $D$ grows from 4 to 32. This is consistent with competition between similar plane channels; a weight norm alone cannot establish which inputs cause the RD change. More planes give a finer geometric discretization in Proposition~\ref{prop:depth} but also give the same output channels more inputs to distinguish. Four planes are a compact deployment candidate; the quality and success results for the nine-rate deployed model use 32.

The calibration perturbations in Fig.~\ref{fig:ablation}c preserve a saving: $-23.71$, $-23.29$ and $-21.81\%$ at $1^\circ$/5\,mm, $2^\circ$/10\,mm and $5^\circ$/20\,mm. A time mismatch is different. Pose jitter yields $+3.69\%$ versus clean HIST and an observation--state offset of one frame yields $+43.78\%$. Sharing an imperfect but fixed transform keeps both predictors synchronized; feeding them a pose corresponding to a different image can make the current warp systematically wrong.

\subsection{Closed-Loop Success across Rates, Resolutions and Policies}\label{sec:cl}
At $256\times256$, KHC-LR reaches 72.6\%, 86.0\% and 88.9\% on OpenVLA-OFT at 6, 8.3 and 10.7\,kbps budgets. Its measured total rates are 5.813, 7.690 and 9.629\,kbps. Relative to HIST-LR the paired gains are 23.625, 5.750 and 3.250 percentage points (pp). The largest gain occurs at the lowest budget. Figure~\ref{fig:teaser} plots x265 at its actual, above-budget rate.

\input{tables/t_cl}
\input{tables/t_lr_cl}

When the two codecs use the \emph{same} $128\times128$ transmit resolution, the 4\,kbps comparison is 386/800 versus 145/800 successes (48.25\% versus 18.13\%) at 3.851 versus 3.919\,kbps. The difference is $+30.13$\,pp, cluster CI $[22.00,38.75]$, Holm-adjusted $p=5.62\times10^{-62}$. At 6\,kbps the corresponding counts are 497/800 versus 289/800 (62.13\% versus 36.13\%) at 5.531 versus 5.735\,kbps: $+26.00$\,pp, CI $[18.63,33.88]$, adjusted $p=5.55\times10^{-50}$. Geometry therefore retains a success advantage at matched resolution; Table~\ref{tab:suites} gives the suite-level differences.

With the independently trained flow-matching policy $\pi_{0.5}$, KHC-LR succeeds in 737/800 trials and HIST-LR in 693/800 at measured rates of 12.588 and 12.594\,kbps. This is 92.13\% versus 86.63\%, a difference of $+5.50$\,pp (CI $[1.00,10.75]$, adjusted $p=7.29\times10^{-5}$). There are 82 KHC-only and 38 HIST-only successes. All four suites favor KHC-LR.

\subsection{Separating Coding from Skipped-Frame Synthesis}\label{sec:channels}
The base-model study uses 13 configurations and 1,600 episodes per configuration. Three measured arms form a sequence of controlled changes: HIST with identity SKIP, KHC coding with identity SKIP, and KHC coding with kinematic SKIP. At 6\,kbps the first change adds 7.63\,pp; the second raises success from 30.375\% to 43.125\%, exactly $+12.75$\,pp. The total difference is about 20.4\,pp. Even a skip-enabled history encoder cannot update the apparent position of the arm when it repeats the previous raster. KHC can. The size of this benefit declines as fewer frames are skipped at higher budgets.

\widefig{1}{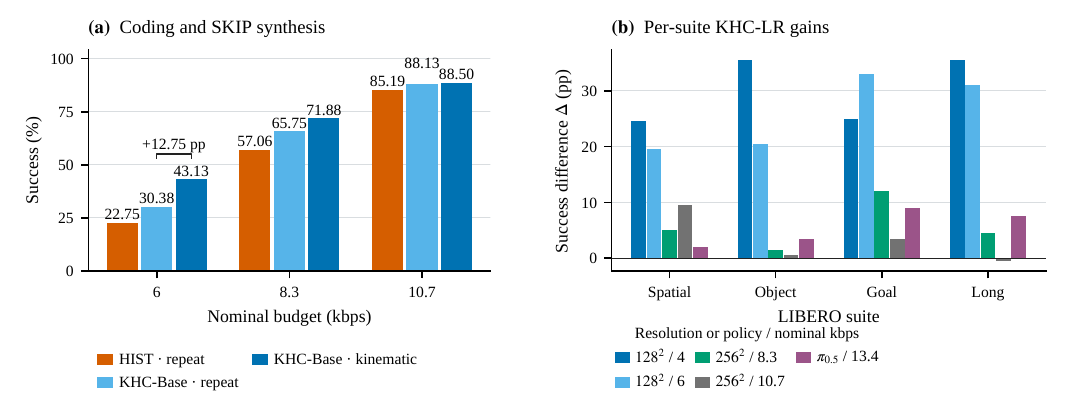}{Three observed base-model configurations separate changes to predictive coding from changes to SKIP synthesis; the identity-to-kinematic gain at 6\,kbps is $+12.75$ percentage points. The suite breakdown retains each comparison's original population and budget.}{fig:decomp}
\widefig{1}{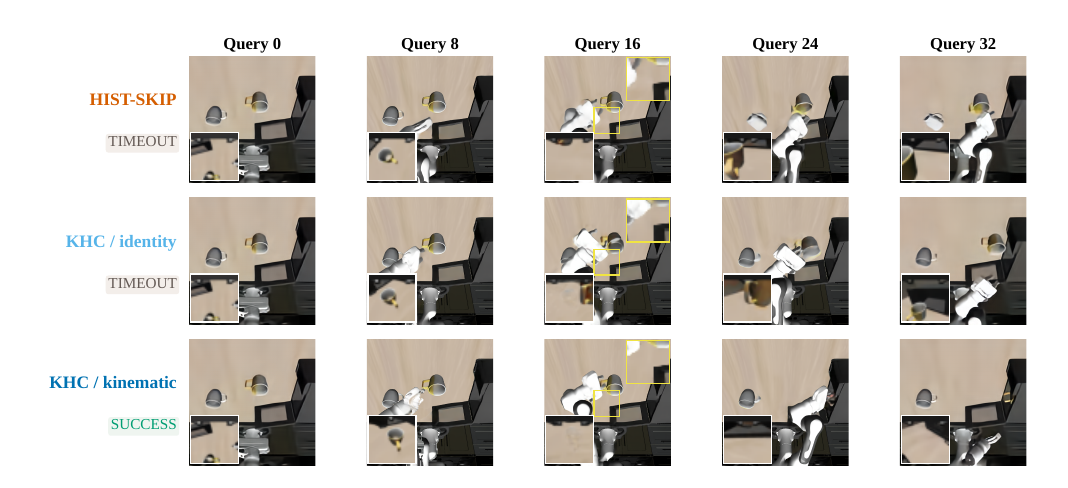}{Paired example on LIBERO-Long task 09, initial state 01 (zero-based). Each row shows the fixed view actually supplied to the policy; the white inset is the simultaneous wrist view. At query 16 the yellow inset enlarges the boxed interaction region. Columns are query indices. The kinematic arm succeeds and the other two time out. Each row then follows its own trajectory.}{fig:qualcl}
\widefig{1}{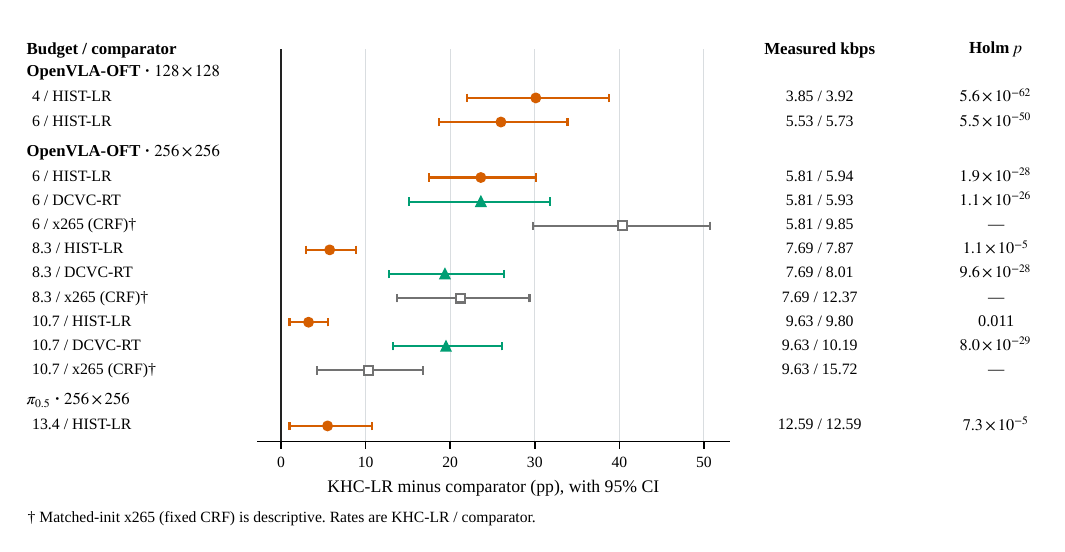}{Paired success differences for the matched-resolution, full-resolution and second-policy experiments. Intervals resample tasks; adjusted probabilities use the exact paired test and the comparison's original Holm family.}{fig:forest}

Figure~\ref{fig:qualcl} follows one initial condition, while Figs.~\ref{fig:decomp} and~\ref{fig:forest} summarize paired populations. A rate-shift model explains why gains concentrate near the steep region; it does not quantitatively predict every arm. Using the separately measured pilot curve, the PSNR-based model has 18.00\,pp mean absolute prediction error on the identity-SKIP arm over the three budgets; replacing PSNR by action deviation yields 19.71\,pp. These errors quantify the limits of predicting closed-loop success from offline distortion alone.

\subsection{Transfer to Two Real Robots}\label{sec:transfer}
On DROID, adapted KHC has PSNR BD-rate $-4.71\%$ across all three cameras and $-11.16\%$ on the moving wrist (95\% CI $[-13.22,-8.96]\%$; Fig.~\ref{fig:transfer}a). The larger wrist saving is consistent with camera-motion parallax: the history reference already captures much of the temporal redundancy in the two nearly stationary exterior views. Reference KHC averages 68.31\,ms for encoding and decoding per camera frame versus 36.20\,ms for HIST ($+88.7\%$, I and P frames included).

\input{tables/t_lr_bd}

On the UR5, the networks receive no target-domain updates. KHC-LR requires $+24.91\%$ and $+40.31\%$ more bits than HIST-LR at equal pooled PSNR for strides two and four. The wrist-only changes are $+27.41\%$ and $+38.53\%$, respectively (Table~\ref{tab:transfer}). Median and 90th-percentile relative hand rotations on the calibration split are $0.73^\circ$ and $4.48^\circ$. The fitted horizontal focal length decreases from 512 to 234 pixels while the photometric loss decreases by 8.69\%. Better photometric fit is not evidence of better metric calibration. Different image statistics, camera coverage and the lack of link rendering also distinguish this setting from the adapted Franka test; these two datasets do not isolate any one of those effects.

\subsection{Transmit Time and the Action-Horizon Boundary}\label{sec:systems}
For the deployed nine-rate weights, sender and receiver reconstructions match bit for bit on all 79,668 test frame--rate records within each implementation. Across implementations the fast path has full-range PSNR BD-rate $+0.00029\%$ (95\% CI $[-0.00110,0.00181]\%$) and MS-SSIM BD-rate $+0.00105\%$ relative to reference (Supplementary Material, Sec.~S-II). On an NVIDIA GeForce RTX 4090 GPU, the matched 100k, $D=32$ profiles give median dual-camera transmit times of 229.4 and 114.5\,ms for reference and fast with four host threads ($-50.1\%$), or 233.0 and 113.5\,ms with two ($-51.3\%$). With four threads the four-plane fast model takes 109.8\,ms and peaks at 455\,MiB (Table~\ref{tab:systems}).

\input{tables/t_features}
\widefig{1}{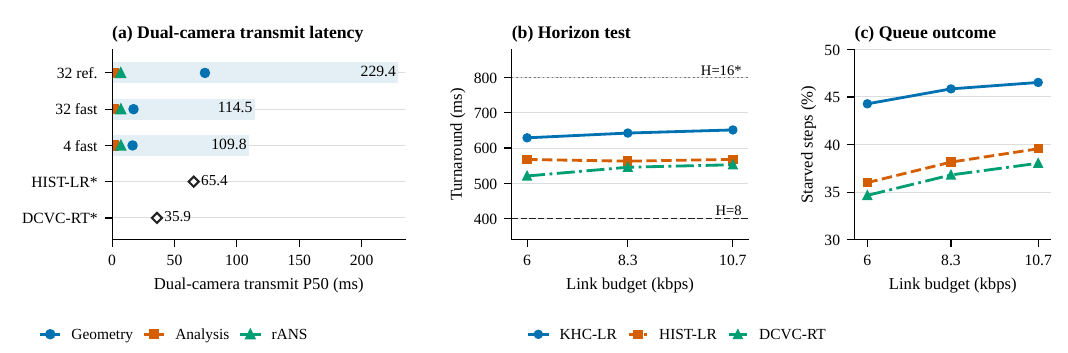}{(a) Bars time the whole dual-camera transmit call, including lower-bound screening, robot-layer rendering, latent analysis, entropy coding and local reference reconstruction. Markers show three nested spans: geometry, analysis and rANS encoding. Diamonds show natural P/P transmit medians. (b) Query turnaround versus the tested $H=8$ and illustrative $H=16$ horizons. (c) Fraction of control steps waiting for an action.}{fig:systems}

For a serial link of rate $R$ bits/s and a packet of $B$ bytes, one query occupies
\begin{equation}
 L_{\rm loop}=L_{\rm enc}+8B/R+L_{\rm dec}+L_\pi+
 L_{\rm RTT}.
 \label{eq:latency}
\end{equation}
At 20\,Hz an eight-action chunk spans $H\Delta t=400$\,ms. The measured policy inference median is 130.4\,ms; RTT adds 40\,ms before accounting for packet size and codec computation. Across nine ordinary method--rate cells, median-packet P/P turnaround spans 521--651\,ms, or 11--14 control steps. Higher nominal rates do not guarantee a shorter query: the controller also admits larger packets.

In 4,800 established-session stop-and-wait rollouts, each ordinary configuration of DCVC-RT, HIST-LR and KHC-LR at 6, 8.3 and 10.7\,kbps solves fewer than 2\% of tasks. Between 34.7\% and 46.5\% of control steps lack a new action, and the step-weighted mean age of information exceeds one second. The experiment includes separate loss and nonstaggered conditions (Supplementary Material, Sec.~S-III). These results identify a scheduling limit: without overlap or queued actions, sustained coverage requires $L_{\rm loop}\le H\Delta t$. Meeting the inequality is not sufficient for task success, but breaking it repeatedly leaves the controller without commands even if the reconstructed scene is adequate.

%% file: tables/t_bd_main.tex
\begin{table*}[t]
\centering
\caption{Pooled-Camera BD-Rate (\%) at Stride 8 on the 200 LIBERO Test Trajectories}
\label{tab:bdmain}
\footnotesize
\setlength{\tabcolsep}{3pt}
\begin{tabular}{@{}l@{\qquad}c@{\qquad}rrrrrr@{}}
\toprule
\multicolumn{1}{c}{Anchor} & Anchor bpp & \multicolumn{2}{c}{PSNR} & MS-SSIM & LPIPS & \multicolumn{2}{c}{Action $L_1$} \\
 &  & Full & Low & Full & Full & Full & Low \\
\midrule
\multicolumn{8}{@{}l}{\textbf{KHC-LR}} \\
HIST-LR & 0.0162--0.1358 & -27.61 & -28.74 & -23.31 & -28.67 & -29.68 & -31.09 \\
HIST (200k) & 0.0278--0.1359 & -28.25 & -29.08 & -19.12 & -26.62 & -28.59 & -30.05 \\
VTM-23 (4:2:0) & 0.0151--0.2573 & +16.13 & +2.05 & +6.63 & -4.91 & \textbf{-5.10} & \textbf{-12.04} \\
VTM-23 (4:4:4) & 0.0101--0.1707 & +22.66 & +5.97 & +9.21 & +2.93 & -1.95 & -8.68 \\
x265 (4:4:4) & 0.0568--0.4192 & -51.70 & N/O & -53.70 & -44.20 & -50.67 & N/O \\
x264 (4:2:0) & 0.0519--0.3756 & -59.60 & N/O & -55.63 & -56.22 & \textbf{-52.44} & N/O \\
SVT-AV1 & 0.0296--0.4700 & -45.64 & -51.36 & -45.45 & -51.61 & \textbf{-56.15} & \textbf{-59.11} \\
DCVC-RT & 0.0069--0.1721 & +0.11 & -5.72 & +3.43 & -24.74 & \textbf{-30.14} & \textbf{-34.16} \\
DCVC-DC & 0.0614--0.1724 & N/O & N/O & +86.90 & +40.46 & +22.48 & N/O \\
KHC-Base & 0.0216--0.1295 & -5.51 & -5.13 & -0.41 & -1.47 & -2.50 & -2.44 \\
\midrule
\multicolumn{8}{@{}l}{\textbf{KHC-Base}} \\
HIST-LR & 0.0162--0.1358 & -21.80 & -22.91 & -18.04 & -23.91 & -24.54 & -26.27 \\
HIST (200k) & 0.0278--0.1359 & -23.91 & -24.89 & -18.19 & -24.55 & -26.20 & -27.59 \\
VTM-23 (4:2:0) & 0.0151--0.2573 & +29.01 & +11.19 & +16.66 & +14.05 & +6.45 & -5.19 \\
VTM-23 (4:4:4) & 0.0101--0.1707 & +42.08 & +18.67 & +25.34 & +25.25 & +11.39 & -0.65 \\
x265 (4:4:4) & 0.0568--0.4192 & -48.71 & N/O & -51.82 & -40.12 & -48.81 & N/O \\
x264 (4:2:0) & 0.0519--0.3756 & -54.61 & N/O & -50.73 & -44.53 & -47.96 & N/O \\
SVT-AV1 & 0.0296--0.4700 & -39.69 & -47.16 & -40.79 & -41.13 & -50.57 & -56.06 \\
DCVC-RT & 0.0069--0.1721 & +7.52 & -2.23 & +9.29 & -13.62 & -23.19 & -31.14 \\
DCVC-DC & 0.0614--0.1724 & N/O & N/O & +87.51 & +42.04 & +26.11 & N/O \\
\bottomrule
\end{tabular}
\par\smallskip
\begin{minipage}{\textwidth}
\footnotesize
Negative values indicate fewer bits for the listed KHC model at equal quality. Full uses the measured common quality interval. Low caps that interval at the test curve's quality at $0.041$ bpp and requires observed low-rate support from both arms. N/O indicates absent support. Action uses the joint camera pair. Sec.~S-II reports coverage and 95\% intervals from 1,000 paired trajectory bootstrap draws. The step-matched 100k comparison is in Table~\ref{tab:ablation}.
\end{minipage}
\end{table*}

%% file: tables/t_ablation.tex
\begin{table}[t]
\centering
\caption{Step-Matched 100k Ablations Against HIST-100k}
\label{tab:ablation}
\footnotesize
\setlength{\tabcolsep}{3pt}
\begin{tabular}{@{}lrrrr@{}}
\toprule
Variant & \shortstack{BD-rate\\(\%)} & 95\% CI & \shortstack{$\Delta$ vs.\\full (pp)} & \shortstack{Plane slot\\share (\%)} \\
\midrule
Full KHC ($D=32$) & -25.67 & [-26.58, -24.72] & 0.00 & 80.2 \\
w/o robot warp & -6.05 & [-6.77, -5.31] & +19.62 & \textemdash \\
w/o plane sweep & -19.10 & [-20.11, -18.15] & +6.57 & \textemdash \\
w/o background & -21.79 & [-23.05, -20.48] & +3.88 & \textemdash \\
w/o EE plane & -22.14 & [-23.54, -20.74] & +3.53 & \textemdash \\
$D=4$ & -27.68 & [-28.60, -26.84] & -2.01 & 44.6 \\
$D=8$ & -23.57 & [-24.60, -22.62] & +2.10 & 59.4 \\
$D=16$ & -23.38 & [-24.57, -22.16] & +2.29 & 72.0 \\
Oracle depth & -39.16 & [-40.15, -38.23] & -13.49 & \textemdash \\
\bottomrule
\end{tabular}
\par\smallskip
\begin{minipage}{\columnwidth}
\footnotesize
PSNR, both cameras, 200 trajectories and 1,000 paired trajectory bootstrap draws. $\Delta$ subtracts the full model's point estimate from each BD-rate against HIST-100k. Plane slot share is the $L_1$ norm of depth-plane channels divided by the total $L_1$ norm across all $4J$ hypothesis-slot channels (share of full stem input including masks and broadcast channels is 26.4\%, 40.1\%, 55.3\%, and 67.5\% for $D=4,8,16,32$; Sec.~S-IV).
\end{minipage}
\end{table}

%% file: tables/t_cl.tex
\begin{table*}[t]
\centering
\caption{Matched Closed-Loop Success Across Uplink Budgets, Resolutions, and Policies}
\label{tab:cl}
\footnotesize
\setlength{\tabcolsep}{3pt}
\begin{tabular}{@{}lllrrlr@{}}
\toprule
Policy / pixels & Budget & Comparator & KHC-LR & Comparator & Paired $\Delta$ (pp) [95\% CI] & Holm $p$ \\
\midrule
OFT $256^2$ & 6 & HIST-LR & 72.63 [5.813] & 49.00 [5.945] & +23.63 [17.50, 30.13] & $1.9\times10^{-28}$ \\
 & 6 & DCVC-RT & 72.63 [5.813] & 49.00 [5.935] & +23.63 [15.13, 31.75] & $1.1\times10^{-26}$ \\
 & 6 & x265 (CRF)$^\dagger$ & 72.63 [5.813] & 32.25 [9.855] & +40.38 [29.75, 50.75] & \textemdash \\
 & 8.3 & HIST-LR & 86.00 [7.690] & 80.25 [7.872] & +5.75 [3.00, 8.88] & $1.1\times10^{-5}$ \\
 & 8.3 & DCVC-RT & 86.00 [7.690] & 66.63 [8.010] & +19.38 [12.75, 26.38] & $9.6\times10^{-28}$ \\
 & 8.3 & x265 (CRF)$^\dagger$ & 86.00 [7.690] & 64.75 [12.370] & +21.25 [13.75, 29.38] & \textemdash \\
 & 10.7 & HIST-LR & 88.88 [9.629] & 85.63 [9.804] & +3.25 [1.00, 5.50] & $0.011$ \\
 & 10.7 & DCVC-RT & 88.88 [9.629] & 69.38 [10.188] & +19.50 [13.25, 26.13] & $8.0\times10^{-29}$ \\
 & 10.7 & x265 (CRF)$^\dagger$ & 88.88 [9.629] & 78.50 [15.718] & +10.38 [4.25, 16.75] & \textemdash \\
\addlinespace
OFT $128^2$ & 4 & HIST-LR & 48.25 [3.851] & 18.13 [3.919] & +30.13 [22.00, 38.75] & $5.6\times10^{-62}$ \\
 & 6 & HIST-LR & 62.13 [5.531] & 36.13 [5.735] & +26.00 [18.63, 33.88] & $5.5\times10^{-50}$ \\
\addlinespace
$\pi_{0.5}$ $256^2$ & 13.4 & HIST-LR & 92.13 [12.588] & 86.63 [12.594] & +5.50 [1.00, 10.75] & $7.3\times10^{-5}$ \\
\bottomrule
\end{tabular}
\par\smallskip
\begin{minipage}{\textwidth}
\footnotesize
Budget and bracketed rates are in kbps. Success is in percent (800 pairs per row). Intervals in the $\Delta$ column use 10,000 suite--task cluster draws. Rates are episode means. $^\dagger$Fixed-CRF x265 rows are descriptive at the displayed measured rates and have no registered $p$ value. OFT is OpenVLA-OFT.
\end{minipage}
\end{table*}

%% file: tables/t_lr_cl.tex
\begin{table*}[t]
\centering
\caption{Per-Suite Success Differences and Closed-Loop Channel Controls}
\label{tab:suites}
\footnotesize
\setlength{\tabcolsep}{3pt}
\begin{minipage}[t]{0.48\textwidth}\centering
\textbf{Suite-level KHC-LR gains (pp)}\par\smallskip
\begin{tabular}{@{}llrrrr@{}}
\toprule
Policy & kbps & Spatial & Object & Goal & Long \\
\midrule
OFT $128^2$ & 4 & +24.5 & +35.5 & +25.0 & +35.5 \\
OFT $128^2$ & 6 & +19.5 & +20.5 & +33.0 & +31.0 \\
OFT $256^2$ & 8.3 & +5.0 & +1.5 & +12.0 & +4.5 \\
OFT $256^2$ & 10.7 & +9.5 & +0.5 & +3.5 & -0.5 \\
$\pi_{0.5}$ & 13.4 & +2.0 & +3.5 & +9.0 & +7.5 \\
\bottomrule
\end{tabular}
\end{minipage}\hfill\begin{minipage}[t]{0.49\textwidth}\centering
\textbf{Coding and SKIP synthesis (success \%)}\par\smallskip
\begin{tabular}{@{}lrrr@{}}
\toprule
Arm or contrast & 6 & 8.3 & 10.7 \\
\midrule
HIST (repeat SKIP) & 22.75 & 57.06 & 85.19 \\
KHC-Base (repeat SKIP) & 30.38 & 65.75 & 88.13 \\
KHC-Base (kin. SKIP) & 43.13 & 71.88 & 88.50 \\
\midrule
Kinematic P coding (pp) & +7.63 & +8.69 & +2.94 \\
Kinematic SKIP (pp) & +12.75 & +6.13 & +0.38 \\
Combined difference (pp) & +20.38 & +14.81 & +3.31 \\
\bottomrule
\end{tabular}
\end{minipage}
\par\smallskip
\begin{minipage}{\textwidth}
\footnotesize
Left: KHC-LR minus HIST-LR over 200 matched trials per suite and budget. Right: 1,600 trials per arm and budget from the 20,800-episode main cohort. Repeat SKIP displays the previous reconstruction. Kinematic SKIP displays the motion-conditioned prediction. Sec.~S-III reports the paired tests for these three arms.
\end{minipage}
\end{table*}

%% file: tables/t_lr_bd.tex
\begin{table*}[t]
\centering
\caption{Real-Robot Transfer and Region-Wise Symbol Information}
\label{tab:transfer}
\footnotesize
\setlength{\tabcolsep}{3pt}
\begin{minipage}[t]{0.50\textwidth}\centering
\textbf{Transfer at matched PSNR (\%)}\par\smallskip
\begin{tabular}{@{}llrrr@{}}
\toprule
Evaluation & View & Stride & BD-rate & 95\% CI \\
\midrule
LIBERO clean & Pooled & 8 & -23.91 & [-25.05, -22.81] \\
DROID adapted & Wrist & 4 & -11.16 & [-13.22, -8.96] \\
DROID adapted & Pooled & 4 & -4.71 & [-5.74, -3.72] \\
UR5 zero shot & Wrist & 2 & +27.41 & [23.38, 30.99] \\
UR5 zero shot & Pooled & 2 & +24.91 & [22.69, 27.32] \\
UR5 zero shot & Wrist & 4 & +38.53 & [33.70, 43.22] \\
UR5 zero shot & Pooled & 4 & +40.31 & [36.80, 43.94] \\
\bottomrule
\end{tabular}
\end{minipage}\hfill\begin{minipage}[t]{0.46\textwidth}\centering
\textbf{Regional information savings (\%)}\par\smallskip
\begin{tabular}{@{}lrrrr@{}}
\toprule
View & Stride & Robot & Static & Novo area \\
\midrule
Fixed & 1 & 47.71 & 8.55 & 0.58 \\
Fixed & 4 & 50.83 & 11.68 & 1.89 \\
Fixed & 16 & 40.60 & 14.67 & 5.44 \\
Wrist & 1 & 39.83 & 6.83 & 2.97 \\
Wrist & 4 & 40.47 & 2.40 & 10.80 \\
Wrist & 16 & 40.47 & -5.58 & 30.99 \\
\bottomrule
\end{tabular}
\end{minipage}
\par\smallskip
\begin{minipage}{\textwidth}
\footnotesize
LIBERO uses simulator geometry (200 trajectories). DROID uses wrist self-calibration and 100{,}000 adaptation steps (30 held-out trajectories). UR5 uses 80 held-out trajectories with frozen wrist geometry, median rotation 0.73$^\circ$, and $f_x$ 512 $\to$ 233.7 px. The fixed and wrist regional rows pool $\alpha\in\{0,3,6\}$ and four motion quartiles over 80 trajectories. Robot and Static report savings in estimated symbol information. Novo area is the newly exposed fraction. Calibration and interval details are in Sec.~S-IV.
\end{minipage}
\end{table*}

%% file: tables/t_features.tex
\begin{table*}[t]
\centering
\caption{Codec Timing Scopes and the Physical Action-Horizon Boundary}
\label{tab:systems}
\footnotesize
\setlength{\tabcolsep}{3pt}
\begin{minipage}[t]{0.48\textwidth}\centering
\textbf{Median transmit and receive times (ms)}\par\smallskip
\begin{tabular}{@{}l@{\enspace}rr@{\enspace}rrr@{}}
\toprule
Codec & \multicolumn{2}{c}{Closed-loop P/P} & \multicolumn{3}{c@{}}{Instrumented dual-camera} \\
 & Tx & Rx & Tx 4 thr. & Tx 2 thr. & Peak MiB \\
\midrule
DCVC-RT & 35.9 & 21.1 & \multicolumn{3}{c}{\textemdash} \\
HIST-LR & 65.4 & 39.7 & \multicolumn{3}{c}{\textemdash} \\
KHC $D=32$, ref. & 133.2 & 73.3 & 229.4 & 233.0 & 573 \\
KHC $D=32$, fast & \multicolumn{2}{c}{\textemdash} & 114.5 & 113.5 & 576 \\
KHC $D=4$, fast & \multicolumn{2}{c}{\textemdash} & 109.8 & 112.4 & 455 \\
\bottomrule
\end{tabular}
\end{minipage}\hfill\begin{minipage}[t]{0.49\textwidth}\centering
\textbf{Physical loop (zero loss, 400 per row)}\par\smallskip
\begin{tabular}{@{}lrrrrr@{}}
\toprule
Codec & kbps & $L$ (ms) & Starve & AoI (s) & Rate share \\
\midrule
KHC-LR & 6 & 629.0 & 44.3 & 1.18 & 46.1 \\
HIST-LR & 6 & 567.5 & 36.0 & 1.03 & 53.7 \\
DCVC-RT & 6 & 520.8 & 34.7 & 1.02 & 60.1 \\
KHC-LR & 8.3 & 642.5 & 45.8 & 1.17 & 48.0 \\
HIST-LR & 8.3 & 562.8 & 38.2 & 1.03 & 54.9 \\
DCVC-RT & 8.3 & 545.5 & 36.8 & 1.05 & 61.8 \\
KHC-LR & 10.7 & 651.4 & 46.5 & 1.17 & 49.4 \\
HIST-LR & 10.7 & 567.9 & 39.6 & 1.04 & 56.0 \\
DCVC-RT & 10.7 & 552.7 & 38.0 & 1.06 & 62.6 \\
\bottomrule
\end{tabular}
\end{minipage}
\par\smallskip
\begin{minipage}{\textwidth}
\footnotesize
Left: NVIDIA GeForce RTX 4090 GPU. Closed-loop P/P times both cameras coding P frames; the KHC row uses deployed KHC-LR. Instrumented transmit uses the matched 100k checkpoints; peak allocated GPU memory is for four host threads. The deployed model's fast/reference PSNR BD-rate is +0.00029\%. VTM-23 and DCVC-DC are offline anchors. Right: $L$ is the frozen normal-P/P round trip at each budget. Action horizon $H\Delta t=400$ ms. Starve is the pooled percentage of control steps without a queued action. Rate share is mean task throughput divided by link rate (\%). Sec.~S-III gives the paired comparisons and packet accounting.
\end{minipage}
\end{table*}

%% file: sec_discussion.tex
\section{Discussion and Conclusion}\label{sec:discussion}
Three decisions govern the robot uplink: which image content the receiver can predict, which residual errors change the policy action, and whether that action can arrive on time.

\emph{Select spatial resolution and temporal rate together.} At approximately 4--6\,kbps, a $128\times128$ KHC stream retains substantially more successful trials than an identically downsampled history stream. At 6--10.7\,kbps the measured $256\times256$ KHC-LR points offer a higher success range; at the common nominal 6\,kbps budget the respective full- and half-resolution KHC-LR rates and success rates are 5.813/5.531\,kbps and 72.6/62.1\%. Above the controller-sensitive range, general-purpose codecs warrant consideration, especially if human viewing quality dominates. For control, action deviation provides a complementary operating-point criterion.

\emph{Pay for geometry only where the receiver can use it.} The compact four-plane model and batched projection are attractive where approximately 110\,ms of dual-camera transmit work is acceptable. Reducing $D$ need not reduce rate efficiency in a fixed-width fusion stem. A low-dimensional correlation or cost volume could separate depth matching from competition for fusion channels. In the wrist view, an independently moving object or a newly visible surface still needs image evidence. A deployable extension could identify these pixels using causal visibility. For cross-robot use, weak wrist rotation makes an independent intrinsic calibration particularly valuable.

\emph{Dimension the control horizon before selecting the codec.} The controller consumes actions at 20\,Hz regardless of network throughput. A stop-and-wait query taking 521--651\,ms cannot reliably replenish an eight-step chunk every 400\,ms. A longer chunk, action prefetch or overlapping computation changes this constraint, but also changes the age of the observations on which actions are based. Real-time chunking~\cite{black2025rtc} and lower-resolution uplinks are therefore different interventions: one changes scheduling, the other payload and distortion. They should be evaluated together under measured latency before making a physical deployment claim.

KHC turns billed robot motion into a predictor available at both endpoints. Where visual errors change the next action, spatially transported history reduces the rate needed for equal action fidelity and improves paired manipulation outcomes across two policies. The loss of that advantage under stale telemetry, weak calibration or an exhausted action horizon identifies what a real deployment must control.